\documentclass[journal]{IEEEtran}

\usepackage{amsmath,amsfonts}
\usepackage{algorithmic}
\usepackage{algorithm}
\usepackage{array}
\usepackage[caption=false,font=normalsize,labelfont=sf,textfont=sf]{subfig}
\usepackage{textcomp}
\usepackage{stfloats}
\usepackage{url}
\usepackage{verbatim}
\usepackage{graphicx}
\usepackage{cite}

\newcommand{\mynote}[1]{}

\newcommand{\mynotex}[1]{}

\IEEEoverridecommandlockouts

\usepackage{graphicx} 

\usepackage{tikz}
\usepackage{textcomp}
\usepackage{hyperref}
\usepackage{lipsum}

\newcommand\copyrighttext{%
  \footnotesize \textcopyright 2026 IEEE. Personal use of this material is permitted.
  Permission from IEEE must be obtained for all other uses, in any current or future
  media, including reprinting/republishing this material for advertising or promotional
  purposes, creating new collective works, for resale or redistribution to servers or
  lists, or reuse of any copyrighted component of this work in other works.
  DOI: \href{https://doi.ieeecomputersociety.org/10.1109/MIC.2026.3695352}{10.1109/MIC.2026.3695352}}
\newcommand\copyrightnotice{%
\begin{tikzpicture}[remember picture,overlay]
\node[anchor=south,yshift=10pt] at (current page.south) {\fbox{\parbox{\dimexpr\textwidth-\fboxsep-\fboxrule\relax}{\copyrighttext}}};
\end{tikzpicture}%
}

\begin{document}

\title{Greening AI Inference with Accuracy and Latency-aware User Incentives}

\author{
\IEEEauthorblockN{Vasilios A. Siris, Adamantia Stamou, George D. Stamoulis, Konstantinos Varsos } \\
\IEEEauthorblockA{Department of Informatics, School of Information Sciences and Technology, \\
Athens University of Economics and Business, Greece} \\
\medskip
\and
\IEEEauthorblockN{Ramin Khalili} \\
\IEEEauthorblockA{Huawei Heisenberg Research Center, Munich, Germany} \\
}

\maketitle
\copyrightnotice
\begin{abstract} 
The widespread use of AI services has raised concerns for its environmental sustainability, towards which recent studies have identified  carbon emissions of AI inference as the major contributor.
This paper introduces a framework for designing AI inference incentives  based on the users’ valuation for inference quality and latency, together with their environmental consciousness, while accounting for the tradeoff between carbon emissions and the two QoE parameters.
Our approach can accommodate different  tradeoffs, that depend on the size and complexity of the AI models and  the  allocation of resources to serve inference requests. 
The incentives can be offered through a practical two-tier service subscription that offers users a discount in exchange for reduced carbon emissions. The discounted service option gives  the AI  provider the flexibility to serve some percentage of inference requests at a lower quality and higher latency during periods of high carbon intensity.
\end{abstract}



\noindent
AI chatbot usage has increased exponentially in recent years, resulting in a significant rise of the energy consumption for providing AI services.
AI inference is currently the dominant source of carbon emissions in deployed AI systems,  contributing up to 90\% of the total systems costs~\cite{D++23}.
The energy intensive processes of inference also create a significant carbon impact: depending on the share of renewable energy, operational emissions can account for up to 70\% of total lifecycle emissions, with embodied emissions comprising the rest~\cite{W++22}.

Although advances in algorithmic efficiency have enabled general purpose AI models to improve their training performance without proportional increases in energy consumption, inference-related energy continues to grow exponentially in newer, cutting edge-models~\cite{D++23}.
The fact that a model is trained once but employed many times further amplifies inference related energy use, a trend that will intensify with the rise of Agentic AI, where autonomous agents execute long horizon tasks with minimal human intervention.
Energy use per inference query also varies widely, from 0.23~Wh for typical prompts to 33~Wh for long ones~\cite{E++25,J++25}, underscoring the need for mechanisms that reduce energy consumption while maintaining acceptable QoE.

Our review of the literature suggests that prior work  treats carbon and resource optimization as a provider-side problem, focusing on the inference accuracy and latency that can be supported by a particular infrastructure and specific AI models. No previous work considers the tradeoff between inference accuracy and latency from the user perspective or the impact this tradeoff has on the design of incentives for users to opt for lower QoE, in exchange for reduced carbon emissions. This user-centric focus is a unique contribution of the proposed framework, allowing us to consider multiple types of users, with different valuations for inference accuracy and latency.


Our specific contribution is two-fold: First, we introduce a new multi-attribute utility framework that quantifies users' valuation for the QoE of AI inference – accuracy and latency – as a function of their environmental consciousness, coupled with a model that captures the dependence of carbon emissions on inference accuracy and latency. 
Our framework is model-agnostic: it does not rely on specific assumptions and it can incorporate alternative models as they become available, making the approach generalizable across platforms and deployments.
%
%
%
Second, based on the introduced utility framework coupled with daily carbon intensity forecasts, we present an approach for constructing a practical two-tier subscription model that provides incentives in the form of subscription discounts, avoiding the complexity and overhead of per-inference incentives. 
Our user-centric design approach provides insight on the appropriate inference accuracy and latency pairs that the provider should offer, based on its user preferences, which
is not possible with a purely provider-focused optimization approach.

\section*{Related work}

Previous work has investigated  mixed-quality AI models to reduce the energy consumption and carbon footprint of AI inference.
Specifically, the work in \cite{L++23} investigates the tradeoffs of using mixed-quality AI models and GPU partitioning to optimize carbon footprint and inference accuracy, while satisfying a  latency constraint. The approach in \cite{L++24} minimizes the carbon footprint subject to a quality constraint with a system-level optimization that considers the probability of selecting different energy consumption and execution time profiles, while \cite{L++25} formulates an Integer Linear Programming (ILP) optimization to minimize the weighted sum of the carbon footprint and the aggregated hardware cost (CPU, GPU, and memory).
The approach in \cite{W++25} minimizes  carbon emissions by balancing the load between service tiers, while satisfying a target percentage of inference requests served by the  higher-quality and higher  emissions tier.

Other work has investigated the cost optimization of inference systems, considering the inference accuracy and latency. 
In particular, the work in \cite{OG20} proposes a framework that maximizes the inference accuracy while satisfying a latency constraint, by executing low accuracy models on a mobile.
The framework  in \cite{S++23} optimizes the weighted sum of the inference accuracy minus a function of the number of CPUs and their load, while satisfying a  latency constraint.
The work in \cite{H++24} investigates a job scheduling and  model selection strategy that satisfies a  quality and  latency constraint, and the work in \cite{G++25} considers an inference pipeline optimization  to maximize  accuracy and minimize cost, while satisfying latency constraints. 
The work in \cite{H++19} exposes service tiers for machine learning cloud services that entail different accuracy tolerance levels, while minimizing service provisioning in terms of response time or server processing time. Finally, the work in \cite{T++24} investigates  model selection that minimizes energy consumption under accuracy and latency constraints.

None of these works incorporate user preferences into decision-making, treating the problem purely from the service provider side.
Additionally, while \cite{W++25} assumes an arbitrary proportion of inference requests served by the higher-quality tier and \cite{H++19} exposes tiers entailing different accuracy and latency tradeoffs,
our framework  defines tiers based on the users' valuation for inference accuracy and latency, taking into account time varying carbon intensities and a target carbon emissions reduction.

\section*{Carbon emissions for AI inference}
\label{sec:carbon_emissions_AI_inference}


An AI provider can offer multiple inference accuracies by utilizing different trained AI models  with a different number of parameters and complexity, hence varying energy consumption and carbon footprint. The values in Table~\ref{tab:carbon_reduction}, which are based on the results in~\cite{L++23}, suggest that the carbon emissions reduction is proportional to the inference accuracy reduction, with the  proportionality factor depending on the specific AI application.
In addition to multiple inference accuracies, 
a provider can offer  multiple inference response times  by sharing GPU resources among multiple  requests. A higher degree of sharing  results in higher response times, but also incurs lower carbon emissions. The bottom part of Table~\ref{tab:carbon_reduction} shows the carbon reduction achieved by providing higher response times~\cite{L++23}.
The experiments in~\cite{L++23} were conducted on a real-world testbed; thus, our evaluation is grounded in empirically observed accuracy–latency–emissions trade-offs.

We  consider  continuous values of the inference accuracy and latency as a function of the  carbon emissions reduction, through  a linear and second-order polynomial interpolation of the values  shown in Table~\ref{tab:carbon_reduction}. Such an approach retains the dependence of the carbon reduction on the inference accuracy and latency, which is based on real measurements, but avoids a  specific quantization that  depends on the  number of AI models and the modes of GPU sharing that a provider supports. This is important since it demonstrates the generality of our framework and its ability to suggest specific inference accuracy and latency values that an AI provider should support based on the characteristics and preferences of its users. Moreover, based on the achievable carbon emissions reduction shown in Table~\ref{tab:carbon_reduction}, our investigations consider a maximum carbon emissions reduction of 80\%; nevertheless, our framework is independent of the specific tradeoffs and dependencies, and  can accommodate other carbon emissions models and experimental results. 



\begin{table}[tb]
\begin{center}
\caption{Carbon emissions reduction for lower inference accuracy and higher latency.}
\label{tab:carbon_reduction}
\begin{tabular}{|c|c|}
\hline \hline Accuracy (normalized)  & Carbon reduction  \\
\hline
1.00 & 0\% \\
0.97 & 20\% \\
0.93 & 40\% \\
0.90 & 60\% \\
0.88 & 80\%\\ 
\hline \hline Latency (normalized)  & Carbon reduction  \\ \hline
1.00 & 0\% \\
1.12 & 19\% \\
1.28 & 26\% \\
\hline
\end{tabular}
\end{center}
\vspace{-0.2in}
\end{table}

\section*{User utility for AI inference and incentives}
\label{sec:utility_incentives}


\mynotex{
\begin{itemize}
\item See if you can/should add the formulas for the two utility curves.
\end{itemize}
}


A utility expresses the user's valuation for a particular QoE, which for AI inference services depends on the inference accuracy and latency.
In practice, the utility of users can be estimated through indirect measurements of  past behavior or direct methods such as user studies, which have  started to appear for AI inference~\cite{M++24, K++25}.

Because a utility for AI inference services is lacking, 
following a standard additive multi-attribute formulation
we introduce the following user utility function that depends on the two QoE parameters:
\begin{equation}
U (a,d) = \lambda \cdot U_{\text{accuracy}} (a) +(1-\lambda) \cdot U_{\text{latency}} (d)  \, ,
\label{eq:utility}
\end{equation}
where $a$ is the normalized inference accuracy, $d$ is the normalized  latency, and $\lambda$ represents the user's valuation of inference accuracy relative to latency: A larger value of $\lambda$ captures a higher valuation of accuracy relative to latency and vise versa.
This additive multi-attribute formulation captures preferential independence, whereby the preferences between the options of one attribute (that is, inference accuracy and latency) do not depend on the value of the other attribute. Moreover, the formulation does not make any assumptions on the accuracy-latency dependency, such as that a higher accuracy always corresponds to a higher latency. 

\begin{figure}[t]
    \centering
    \includegraphics[width=0.95\linewidth]{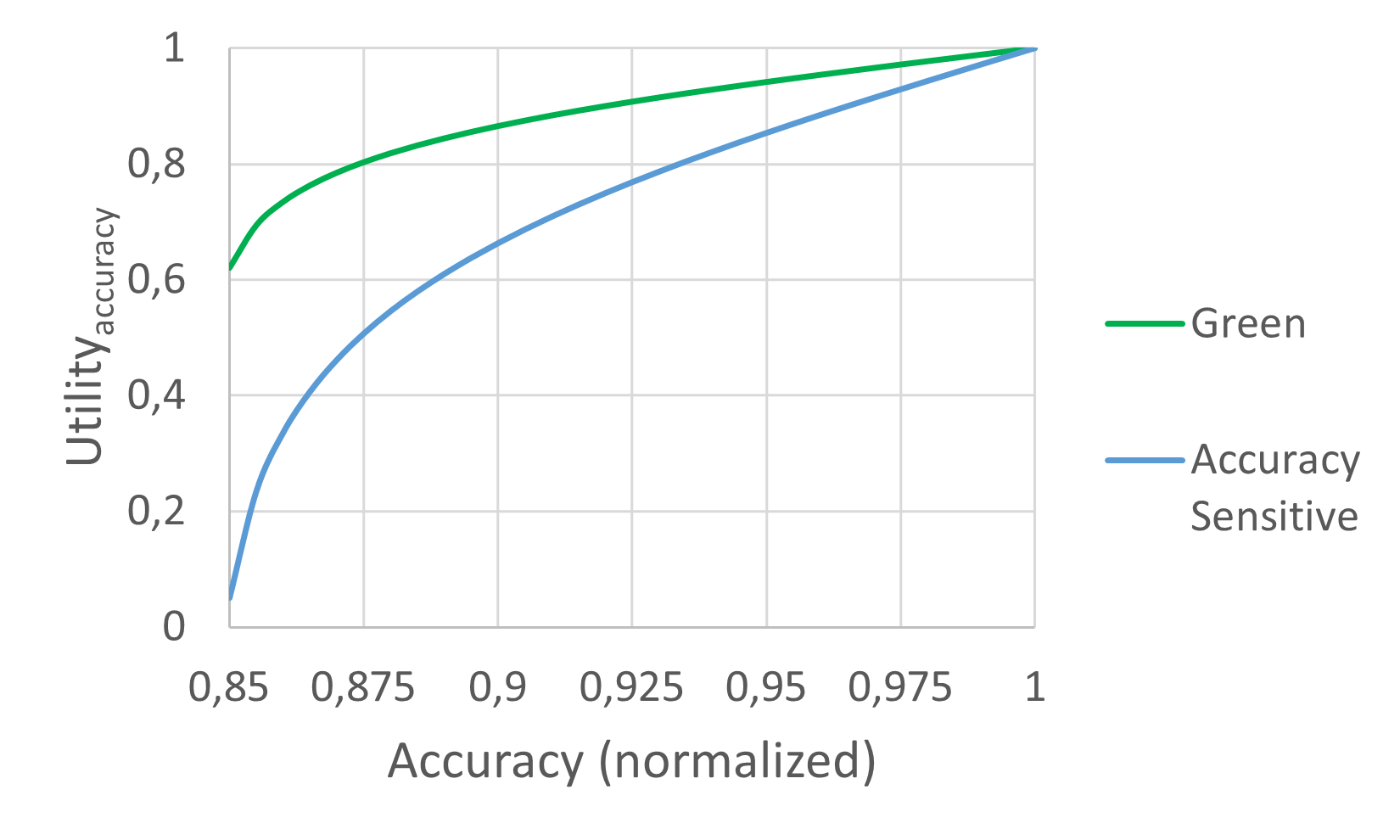}
    \vspace{-0.1in}
    \caption{The inference accuracy is normalized to the accuracy that is achieved with the most complex AI model, which has the largest carbon emissions. A green user achieves the same utility as the accuracy sensitive user, but for a smaller accuracy, except in the case of the highest utility (one).
    }
    \label{fig:utility_accuracy}
    \vspace{-0.1in}    
\end{figure}

Figure~\ref{fig:utility_accuracy} shows the utility of the inference accuracy
as an increasing concave function of  the normalized accuracy; such a shape expresses diminishing marginal returns for increased accuracy, which is a common property of a utility for goods or services.
The specific function we consider is $U_{\text{accuracy}} (a) = 1-\frac{1}{B}\log \left ( \frac{2}{1-6.5(1-a)} -1 \right)$; parameter $B$ captures a user's sensitivity to accuracy and is $B=2$ and $5$ for the accuracy sensitive and green user, respectively; the other fixed parameters of $U_{\text{accuracy}}$ are selected so that the utility covers the whole normalized accuracy  range  of the carbon emissions model in Table~\ref{tab:carbon_reduction}. 
The unique feature of the specific form of the utility (and of the latency utility discussed below) is that the values on the horizontal axis are normalized to the highest accuracy, 
which entails the highest carbon emissions.
Expressing the  utility  as a function of normalized accuracy makes the proposed framework more general, as it can be applied to varying carbon intensities, as discussed below, and is consistent with presenting the QoE of a low-quality tier relative to the highest quality tier.


We consider different types of users with different accuracy sensitivity values because one of our goals is to investigate how they influence the incentives and the carbon emissions reduction. The green or environmental conscious user in Figure~\ref{fig:utility_accuracy} 
achieves the same utility for a smaller accuracy  compared to that of an accuracy-sensitive user. Specifically, a green user achieves utility  0.8 with an accuracy that is approximately 7\% lower than the accuracy that achieves the same utility for an accuracy-sensitive user. These numbers are illustrative and as we  investigate below reflects the green user's  willingness to sacrifice some accuracy for reduced carbon emissions.

\begin{figure}[t]
    \centering
    \includegraphics[width=0.95\linewidth]{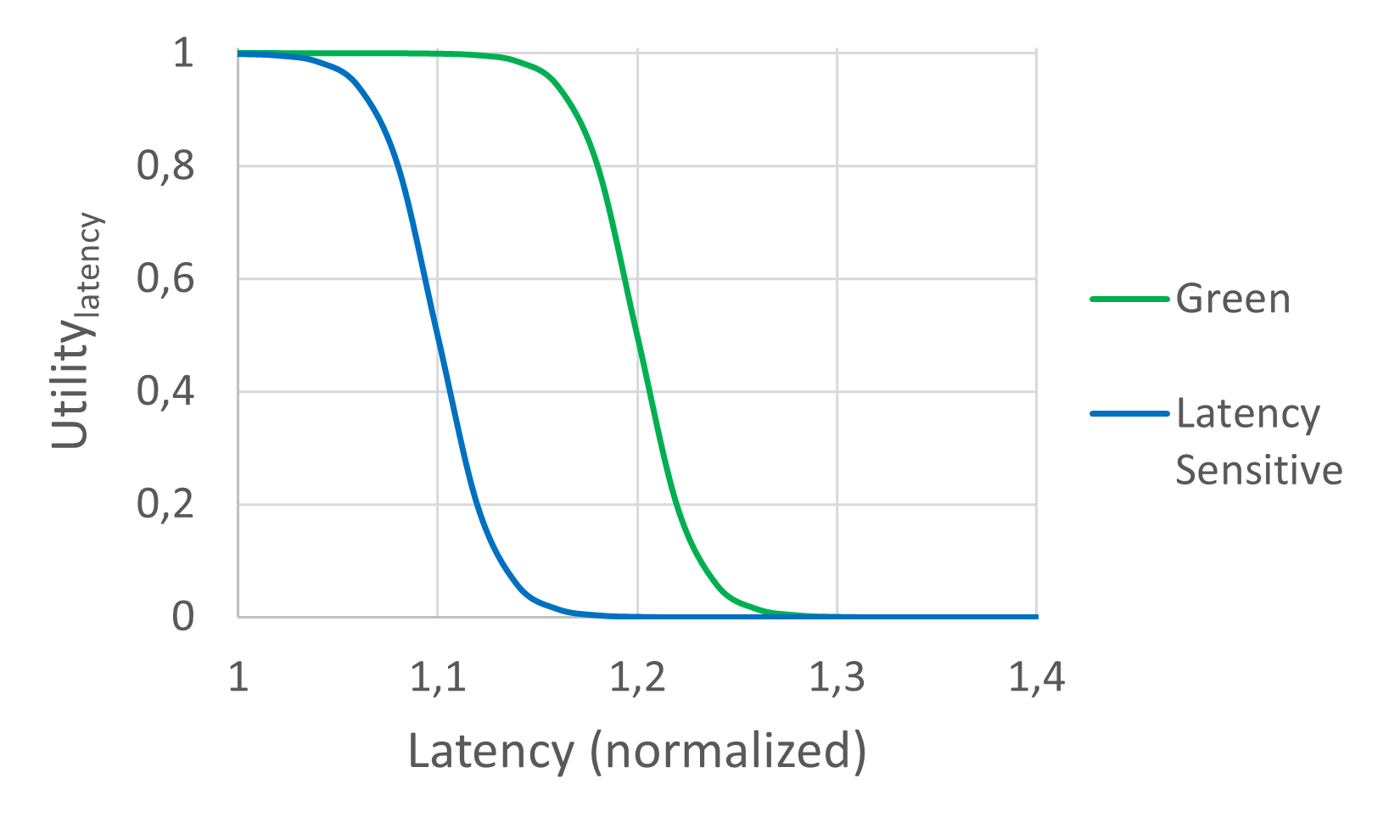}
    \vspace{-0.1in}
    \caption{Initially the utility for inference latency remains high but drops abruptly when the normalized latency increases above some value, which is larger for a green user compared to a latency-sensitive user. }
    \label{fig:utility_latency}
    \vspace{-0.1in}
\end{figure}

Figure~\ref{fig:utility_latency} shows the utility of the inference latency, which approximates a stepwise function. 
The specific function we consider is $U_{\text{latency}} (d) = 1-\frac{1}{1-e^{7-70(d-X)}}$; parameter  $X$ captures the user's tolerance to latency and is  $X=1$ and $1.1$ for the latency sensitive and green user, respectively.
The delay shown on the x-axis of Figure~\ref{fig:utility_latency} is normalized to the minimum response latency, for which the utility is one, while the other fixed parameters of $U_{\text{latency}}$ are selected to cover the whole range of normalized delay  in Table~\ref{tab:carbon_reduction}.
The shape of the utility  is a flipped sigmoid function, which captures empirically observed human perception thresholds for service delays: For latency values  close to one (which corresponds to the lowest latency), the utility is close to its maximum value. When the latency increases above some value, the utility starts to drop fast. The sudden drop of the utility starts at a smaller value for the latency-sensitive user compared to the green user, reflecting the green user's  tolerance to higher latency, hence the willingness to sacrifice latency for reduced carbon emissions. Specifically, a green user achieves a utility of 0.8 for latency that is 9\% higher than the latency that the latency-sensitive user incurs when achieving the same utility (namely, 0.8).


%


Next, consider that a user is presented with an economic incentive to reduce carbon emissions in the form of a linear charge
$C(r)=p \cdot (1-r)$, where $r$ is the carbon emissions reduction and $p$ is the price per unit of normalized carbon emissions. 
Hence, if the user selects the highest inference accuracy and lowest latency, for which the normalized carbon emissions is equal to 1, i.e., $r=0$, the user will incur a charge  $p$.
If  the user  reduces carbon emissions by 20\%, then the charge becomes $0.8 \cdot p$ or equivalently the user receives a discount $0.2 \cdot p$ compared to a high-quality user. 
Parameter $p$ expresses the economic benefit per percentage of carbon emissions reduction and reflects  the incentives provided to users: a higher value of $p$ reflects a higher incentive for users to reduce their requested QoE.
Later we describe a practical tier-based approach in which users are offered a discount when some percentage of their inference requests can be served at lower quality; hence, the carbon emission charge  $p$ in practice is not directly exposed to users.

\mynotex{It is interesting to note that the discount offered to a user that reduce the QoE can combine the carbon emissions charge  with  the usage/energy consumption related cost incurred by the AI inference provider required to serve inference queries. Hence, the offered discount can capture the total usage cost of an inference response, and the corresponding reduced carbon and usage cost if the user reduces the QoE. Decide if this should be mentioned. }

A rational user  facing the charge presented above will seek to maximize his net benefit $\textit{NB}$, given by  the utility minus the charge, i.e.
\begin{equation}
\max_{r} \textit{NB}(r) \,, \text{   where } \textit{NB}(r) = U(r) -C(r) \, .
\label{eq:net_benefit}
\end{equation}
The net benefit optimization is over the carbon emissions reduction  $r$, but  the (rational) user simultaneously selects the combination of the two QoE parameters, namely inference accuracy $a$ and latency $d$, that achieve the maximum net benefit. 
The optimum can be found using a two-dimensional search over $r$ and $a$, since $d$ is determined by $r$ and $a$, and the utility $U(r)$ in (\ref{eq:net_benefit}) can be obtained from (\ref{eq:utility});
this procedure is independent of any particular accuracy-latency tradeoff.
Since the net benefit in (\ref{eq:net_benefit}) is given by the utility minus the charge, the carbon emissions charge (incentive) $p$ is expressed in units of utility.

\mynotex{The optimum is over two variables: $r$ and $a$ or $d$. The optimum was found using a brute force search on a 2D 10x10 discrete value grid, which turned out to be  practical. }

\section*{Carbon reduction for different types of users}
\label{sec:different_user_types}

The investigations of this section answer the following two questions for different types of users that value inference accuracy and latency differently: First, what is the carbon emissions reduction when a user is offered an incentive, expressed by the value of parameter $p$ in the cost function of (\ref{eq:net_benefit}). Second, what is the best combination of inference accuracy and  latency  that achieves this reduction.

We consider four types of users, which differ in their degree of environmental consciousness and their relative valuation of inference accuracy and latency. 
Green users and high-quality users, which are both accuracy-sensitive and latency-sensitive, have  the corresponding utility shown in Figures~\ref{fig:utility_accuracy} and \ref{fig:utility_latency}. Users who equally value the inference accuracy and latency have $\lambda=0.5$ in (\ref{eq:utility}), while users who value inference accuracy nine times more than latency have  $\lambda=0.9$.

\begin{figure}[t]
    \centering
    \vspace{-0.1in}
    \includegraphics[width=1.0\linewidth]{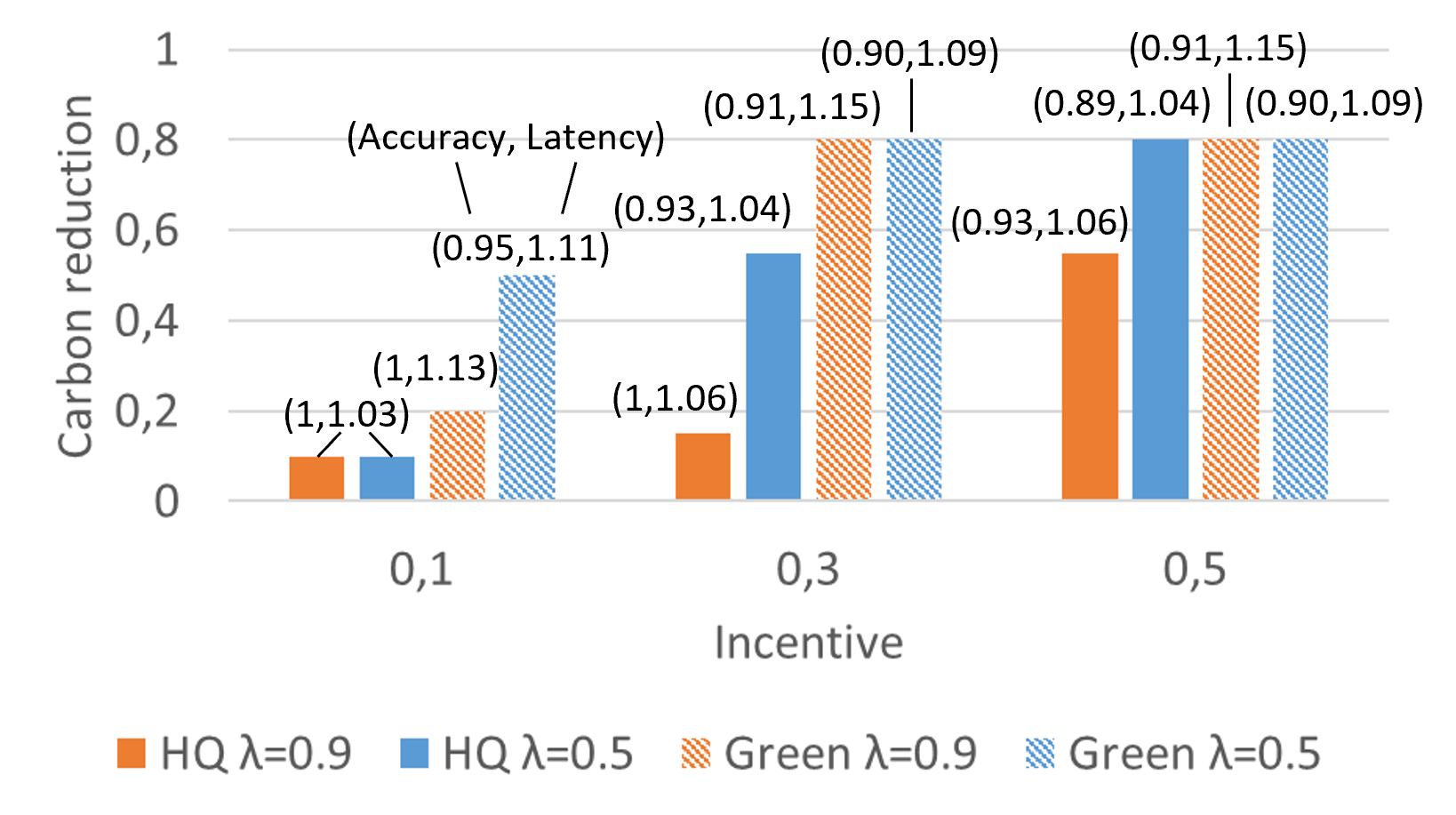}
    \vspace{-0.25in}
    \caption{Carbon reduction and corresponding inference accuracy  and latency  for different incentives $p=0.1, 0.2, 0.3$ and for different user types: high-quality (HQ), Green, users with equal valuation for inference accuracy and latency ($\lambda=0.5$), and users with a higher valuation for accuracy ($\lambda=0.9$).}
    \label{fig:user_types}
\end{figure}

The reduction in carbon emissions and the best corresponding combination of accuracy and inference latency for each type of user when offered an incentive $p=0.1, 0.3, 0.5$ is shown in Figure~\ref{fig:user_types}. 
For the same incentive, a green user typically accepts  lower accuracy and  higher latency compared to a high-quality (HQ) user, which results in lower carbon emissions for the green user. Moreover, users that value  accuracy significantly more than latency (HQ and green users with $\lambda=0.9$ in Figure~\ref{fig:user_types}) accept lower reduction of lower carbon emissions due to the shape of the latency utility curve, which decreases abruptly when the latency is above some value. Finally, the results show that there is a saturation point, $p \geq 0$, which can help providers  effectively select incentives.

The results in Figure~\ref{fig:user_types} also depend on the accuracy-latency-emissions tradeoff,  shown in Table~\ref{tab:carbon_reduction}, in addition to the user valuation for  accuracy and latency. To illustrate this dependence, if a 80\% carbon reduction is achieved for normalized accuracy 0.94, which is higher than the value 0.88 shown in Table~\ref{tab:carbon_reduction}, while retaining the linear dependence of the inference accuracy and emissions reduction, then all users  would tend to select a higher carbon emissions reduction than the reduction shown in Figure~\ref{fig:user_types}, up to the maximum of 80\%. This occurs because with the new accuracy-emissions tradeoff the same reduction of the inference accuracy now achieves a higher emissions reduction. On the other hand, if a 80\% carbon reduction is achieved for normalized accuracy 0.81 rather than 0.88 shown in Table~\ref{tab:carbon_reduction}, then users would tend to select  a lower emissions reduction than the reduction in Figure~\ref{fig:user_types}.
Furthermore, we can relax the linear dependence of the inference accuracy and emissions reduction, in which case the incremental impact of different incentives would be different, since due to non-linearity the same carbon emissions reduction can have a different impact on the inference accuracy.

Regarding the impact of a higher incentive, Figure~\ref{fig:user_types}  shows that, as expected, a higher incentive typically results in the users  accepting a lower inference accuracy and worse latency, in exchange for lower carbon emissions. However, the impact of a higher incentive is lower for a HQ user and a user who values inference accuracy more than latency ($\lambda=0.9$). 
Also, with a higher incentive ($p=0.5$) all user types choose the highest carbon emissions reduction (80\%), except the high-quality user with $\lambda=0.9$.

In addition to the carbon emissions reduction, Figure~\ref{fig:user_types} also shows for each type of user, the best inference accuracy and latency combination that achieves the corresponding carbon emissions reduction. Hence, for a high-quality user and for incentive $p=0.3$, we see that a user that values inference accuracy more than latency ($\lambda=0.9$) selects the highest accuracy (=$1$) but accepts latency 6\% higher than the best. On the other hand, a user that equally values inference accuracy and latency ($\lambda=0.5$) accepts a lower accuracy (93\%) but also a  latency that is 4\% higher than the best.
For this user, the 55\% carbon emissions reduction consists of a 44\% reduction from accepting the lower accuracy and a 11\% reduction from accepting a worst latency.
Another conclusion is that, compared to a high-quality user,
a green user accepts both lower inference accuracy and worst latency, which results in lower carbon emissions. Specifically, for incentive $p=0.3$ the high-quality user selects the best inference quality (=$1$) and latency 6\% higher than the smallest value, whereas the green user accepts a significantly worst inference accuracy and latency: 91\% and 15\%, respectively. These results highlight the importance of educating users on  sustainability and environmental consciousness.


\section*{A practical design for tier-based monthly incentives}
\label{sec:tiers}

In this section we discuss how the  approach presented above can be used to design a two-tier model that can achieve the same carbon emissions reduction by offering users the incentives in the form of a  discounted subscription service, rather than as a per-inference carbon emissions charge.
The discounted service  allows the AI inference provider to serve up to some percentage of inference queries at a lower  accuracy and a higher  latency, which we can estimate.
We assume that these different model tiers are already deployed in the data plane, maintaining warm serving pools, while the control-plane scaling decisions are performed proactively to adjust the available capacity to the predicted workload.


\begin{figure}[tb]
    \centering
    \includegraphics[width=1.0\linewidth]{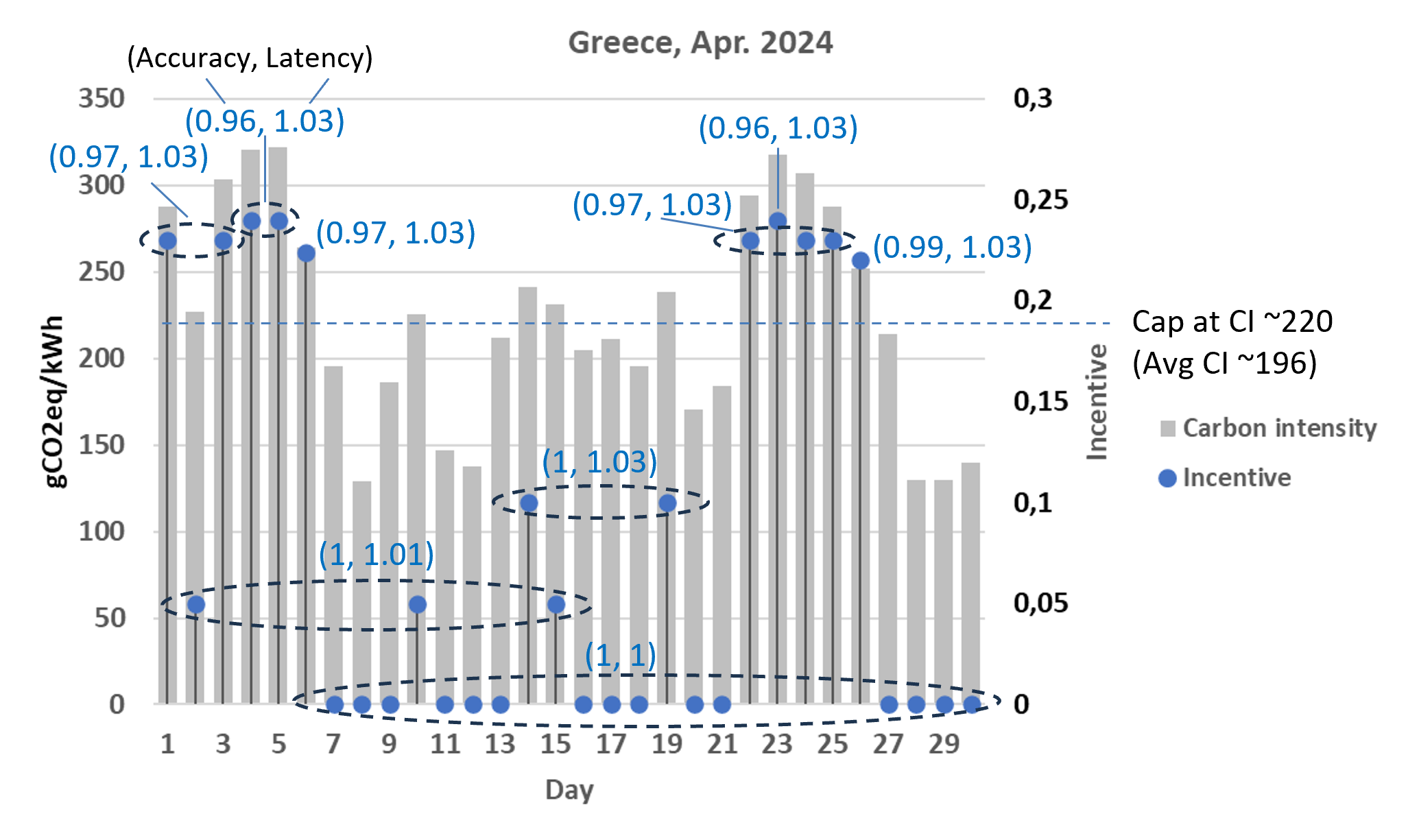}
    \vspace{-0.25in}
    \caption{Incentives for different days of a month to  cap  the total carbon emissions below a threshold (220~gCO$^2$eq/kWh) and the corresponding optimal inference accuracy and latency. When the carbon intensity is below the threshold, no incentives are provided and there is no degradation of the inference QoE.}
    \label{fig:tiers}
    \vspace{-0.1in}
\end{figure}
The light-colored bars in Figure~\ref{fig:tiers} show the carbon intensity for different days of a month (April 2024) in Greece, using data from Electricity Maps\footnote{http://electricitymaps.com}, for a high-quality user with $\lambda=0.5$. 
Assume that an AI inference provider, for each day, wants to cap the carbon emissions below a threshold (220~gCO$^2$eq/kWh). 
The consideration of a  carbon intensity cap is motivated by Emissions Trading Systems (ETSs) that impose a cap on the total emissions, forcing companies to buy allowances once the cap is exceeded~\cite{M-B23}. Indeed, recent studies have shown that most new and planned direct carbon pricing instruments are ETSs~\cite{WorldBank2025}. Providing incentives during periods of high carbon intensity is motivated by regulations such as EU's Corporate Sustainability Reporting Directive (CSRD), that mandates comprehensive disclosure of Scope~2 emissions from purchased energy, which is the dominant source of emissions for data centers. Moreover, regulatory-driven direct carbon pricing, such as ETSs and carbon taxes, directly impose a fixed price per unit of carbon emissions~\cite{WorldBank2025}, thus increasing the operational costs of data centers and offering them monetary incentives to reduce their emissions.


Because the carbon intensity varies for different days, due to the availability of renewable energy, the incentives (shown with dots in~Figure~\ref{fig:tiers}) and  the best  mix of inference accuracy and latency (shown in the parenthesis next to the incentive dots) can vary day-to-day. 
A direct application of our framework that provides different daily incentives to users would be
complicated for users and incur a high interaction and accounting overhead. 
Instead, we propose a different approach where a user can select between two subscriptions: A high-quality subscription where all inference queries are served with the highest accuracy and lowest latency, and a discounted subscription where up to some percentage of the user's queries are served with  lower accuracy and higher latency.
To determine the percentage of lower QoE and the corresponding accuracy and latency pairs of the discounted subscription,  we consider a worst-case approach as follows: 
During days when the carbon intensity is below the target carbon emissions cap, the  provider can answer all  queries with the highest accuracy and lowest latency (high tier); this is the case for 15 out of the 30 days shown in Figure~\ref{fig:tiers}, which corresponds to 50\% of the queries if we assume that  they are equally distributed across all days. For the other 15 days, the provider can promise to answer  queries with an  accuracy that is the lowest value determined by our model and with a latency that is the highest. 
For the results in Figure~\ref{fig:tiers}, the inference accuracy that the provider can promise is at least 96\% while the latency is at most  3\% higher than the best latency (low tier). Finally, the discount of this service option would depend on the incentives that our model determines for each of the 15 days that the carbon intensity is above the threshold and the corresponding carbon emissions reduction when accepting a lower QoE. For the scenario depicted in  Figure~\ref{fig:tiers}, the flexible subscription would have a discount of approximately 27\%.
The insight obtained from Figure~\ref{fig:tiers} allows a provider to estimate, based on the characteristics and preferences of its users, the best inference accuracy and latency pair and the discount of the flexible subscription  to achieve a target carbon intensity threshold.

 

To pre-estimate the discount value the provider would need  a priori knowledge of  the carbon intensity during a month. 
Furthermore, because the carbon intensity profile can vary for different months, the provider can aggregate the  incentives and the inference accuracy and latency across multiple months,  to provide the same discounted subscription option across these months. 
Alternatively, the provider can use a day-ahead forecast of the average carbon intensity to decide the inference accuracy and latency  for that specific day. At the end of the month, the provider computes the discount by averaging across all days.
This alternative functions like an open market where the accuracy and latency of services is not known a priori and the  final discount is determined by how much the users contributed to meeting the carbon emissions target.

Finally, our framework can be adapted to a token-based variant where users consume tokens for high-quality inference requests, which matches  pay-as-you-use AI  services. 
Specifically, depending on a user's subscription, in periods of high carbon intensity the inference accuracy and latency can match what is predicted by our model or high-quality responses would require an amount of tokens that is proportional to the predicted incentives.


\section*{Conclusions}


Ongoing work is adapting the proposed framework to user inference sessions that initially have queries requiring low latency but accepting lower inference accuracy, followed by latter queries requiring higher accuracy but accepting worst latency.
Such an inference query progression is in line with the general to specific progression of web searching but is also a property of AI chatbot sessions~\cite{Y++23}.
Finally, we are investigating the impact of routing inference requests to AI data centers utilizing geolocation carbon intensity information to reduce carbon emissions.

\vspace{-0.1in}
\section*{Acknowledgment}

This work has been developed in part within project EXIGENCE, which is funded from the Smart Networks and Services Joint Undertaking (SNS JU) under the EU Horizon Europe Grant  No 101139120. 

\mynotex{Up to 20 references.}

\bigskip 

\noindent
\textbf{Vasilios A. Siris} is a Professor at the 
Department of Informatics, School of Information Sciences and Technology, Athens University of Economics and Business (AUEB),  Athens, Greece. V.~Siris received his PhD in performance analysis and pricing in broadband networks from the University of Crete, Greece.
His  research interests include efficiency and trust in the Internet and mobile communication systems.
He is a senior member of the IEEE.
Contact him at vsiris@aueb.gr.

\medskip
\noindent
\textbf{Adamantia Stamou}  is a senior researcher at the Department of Informatics, School of Information Sciences and Technology, AUEB.  She is currently engaged in research at the intersection of cutting-edge digital infrastructure and sustainable systems, with a focus on 6G networks, energy efficiency, incentive mechanisms, and blockchain technologies. A. Stamou received her PhD in telecommunications from the National Technical University of Athens. Contact her at stamouad@aueb.gr.

\medskip
\noindent
\textbf{George D. Stamoulis} is a Professor at the Department of Informatics, School of Information Sciences and Technology, AUEB. His main research interests include economic and incentives mechanisms for resource and energy management in 5G/6G networks, clouds, the Internet, and smart energy grids as well as business modeling of these systems. G. Stamoulis received his Ph.D. in optimal routing and performance evaluation in interconnection networks from the Massachusetts Institute of Technology. Contact him at gstamoul@aueb.gr.

\medskip
\noindent
\textbf{Konstantinos Varsos} is a postdoc researcher at the Department of Informatics, School of Information Sciences and Technology, AUEB. K. Varsos received his PhD in game theory from the University of Crete. His research interests include game theory and computational mathematics. Contact him at kvarsos@aueb.gr.

\medskip
\noindent
\textbf{Ramin Khalili} is a principal researcher at Huawei Heisenberg Research Center in Munich. His research centers on resource management and service optimization within large-scale AI and distributed systems. R. Khalili has received his PhD in computer networks and distributed systems from Sorbonne University in Paris. Contact him at ramin.khalili@huawei.com.

\end{document}